\pdfoutput=1
\documentclass[11pt]{article}
\usepackage[final]{acl}
\usepackage{times}
\usepackage{latexsym}
\usepackage[T1]{fontenc}
\usepackage[utf8]{inputenc}
\usepackage{microtype}
\usepackage{inconsolata}
\usepackage{graphicx}
\usepackage{amsmath}
\usepackage{bbm}
\usepackage{amssymb}	
\usepackage{booktabs}
\usepackage{cuted}
\usepackage{epsfig}
\usepackage{caption}
\usepackage{float}
\usepackage{placeins}
\usepackage{color, colortbl}
\usepackage{stfloats}
\usepackage{enumitem}
\usepackage{tabularx}
\usepackage{xstring}
\usepackage{multirow}
\usepackage{multicol}
\usepackage{xspace}
\usepackage{url}
\usepackage{subcaption}
\usepackage{xcolor}
\usepackage[hang,flushmargin]{footmisc}
\usepackage{arydshln}
\usepackage{algorithm}
\usepackage{tcolorbox}
\usepackage{adjustbox}
\usepackage[noend]{algpseudocode}
\usepackage{listings}
\usepackage{enumitem}
\usepackage{fontawesome5}
\usepackage{marvosym}
\tcbuselibrary{breakable} 
\definecolor{Green}{HTML}{66CC99} % 7BB662
\definecolor{Gray}{HTML}{B0B0B0} % A9A9A9
\definecolor{Blue}{HTML}{4A90E2} % 69A7D2
\definecolor{prompt_blue}{HTML}{1f78b4}
\definecolor{prompt_red}{HTML}{d45c43}
\definecolor{plus}{HTML}{0071bc}
\definecolor{minus}{RGB}{153,10,10}
\definecolor{SecondBest}{HTML}{F0FCEB} % E6F9D6
\definecolor{Best}{HTML}{D4F2BA}
\definecolor{cvprblue}{rgb}{0.21,0.49,0.74}
\definecolor{Green}{rgb}{0.2, 0.7, 0.1}
\newcommand{\up}{\bf \fontsize{10}{42} \color{plus}{$\uparrow$}}
\newcommand{\down}{\bf \fontsize{10}{42}\selectfont \color{minus}{$\downarrow$}}
\usepackage[capitalize]{cleveref}
\crefname{section}{Sec.}{Secs.}
\crefname{table}{Table}{Tables}
\crefname{figure}{Fig.}{Figs.}
\usepackage{pifont}
\newcommand{\cmark}{\textcolor{Green}{\ding{51}}}
\newcommand{\xmark}{\textcolor{red}{\ding{55}}}

\newcommand*\rot{\rotatebox{90}}
\newlength\savewidth

\newcolumntype{x}[1]{>{\centering\arraybackslash}p{#1pt}}
\newcolumntype{y}[1]{>{\raggedright\arraybackslash}p{#1pt}}
\newcolumntype{z}[1]{>{\raggedleft\arraybackslash}p{#1pt}}

\newlength\abovesecmargin
\newlength\belowsecmargin
\newlength\abovesubsecmargin
\newlength\belowsubsecmargin
\newlength\abovesubsubsecmargin
\newlength\belowsubsubsecmargin
\newlength\paramargin
\newlength\abovetabcapmargin
\newlength\belowtabcapmargin
\newlength\abovefigcapmargin
\newlength\belowfigcapmargin
\makeatletter
\DeclareRobustCommand\onedot{\futurelet\@let@token\@onedot}
\def\@onedot{\ifx\@let@token.\else.\null\fi\xspace}
\def\eg{\emph{e.g}\onedot}

\makeatother
\title{
    Black-Box Visual Prompt Engineering for Mitigating Object Hallucination in Large Vision Language Models
}

\author{
Sangmin Woo\textsuperscript{$\heartsuit$$\spadesuit$}\thanks{Work done during an internship at Amazon.\\\textsuperscript{\Letter}Corresponding author.} \quad Kang Zhou\textsuperscript{$\heartsuit$} \quad Yun Zhou\textsuperscript{$\heartsuit$} \quad Shuai Wang\textsuperscript{$\heartsuit$} \quad Sheng Guan\textsuperscript{$\heartsuit$} \\ \textbf{Haibo Ding\textsuperscript{$\heartsuit$\scalebox{1.2}{\Letter}}} \quad \textbf{Lin Lee Cheong\textsuperscript{$\heartsuit$}}\\
\textsuperscript{$\heartsuit$}Amazon AWS AI \quad \textsuperscript{$\spadesuit$}KAIST\\
\texttt{\{sangminw,zhoukang,yunzzhou,wshui,shguan,hbding,lcheong\}@amazon.com}
}

\begin{document}
\maketitle

% \begin{abstract}
% Despite their impressive capabilities, Vision-Language Assistants (VLAs) often suffer from object hallucination, severely limiting their reliability in practical applications.
% Visual prompting has shown promise in enhancing visual grounding, but its effectiveness remains inconsistent and not well-understood, particularly in addressing object hallucinations.
% This raises two key questions: \textbf{(Q1)} Can visual prompting mitigate object hallucinations in VLAs? \textbf{(Q2)} Can we systematically learn the optimal visual prompts?
% Prompt engineering for multi-modal domains is still under-explored, especially when treating VLAs as \textit{"black boxes"} with only input-output access.
% To tackle these challenges, we introduce Black-Box Visual Prompt Engineering (\textbf{Black-Box VPE}), a framework designed to identify the optimal visual prompts that elicit improved responses from VLAs without touching themselves.
% Our approach involves maintaining a pool of visual prompts and training a router model to dynamically select the best prompt based on VLA preferences for given input images.
% This black-box approach, which does not require access to model internals, is applicable to both open-source (\eg, LLaVA 1.5, InstructBLIP) and proprietary VLAs (\eg, GPT-4o, Claude-3.0-Sonnet).
% Evaluations on object hallucination benchmarks such as POPE and CHAIR demonstrate that our approach effectively reduces hallucinations and enhances overall model performance.
% \end{abstract}

\begin{abstract} 
Large Vision Language Models (LVLMs) often suffer from object hallucination, which undermines their reliability. Surprisingly, we find that simple object-based visual prompting---overlaying visual cues (\eg, bounding box, circle) on images---can significantly mitigate such hallucination; however, different visual prompts (VPs) vary in effectiveness. To address this, we propose \textbf{Black-Box Visual Prompt Engineering (BBVPE)}, a framework to identify optimal VPs that enhance LVLM responses without needing access to model internals. Our approach employs a pool of candidate VPs and trains a router model to dynamically select the most effective VP for a given input image. This \textit{black-box} approach is model-agnostic, making it applicable to both open-source and proprietary LVLMs. Evaluations on benchmarks such as POPE and CHAIR demonstrate that BBVPE effectively reduces object hallucination.
\end{abstract}

\begin{figure*}[t!]
% \begin{strip}
    \centering
    % \vspace{-28mm}
    \vspace{-4mm}
    \includegraphics[width=0.93\linewidth]{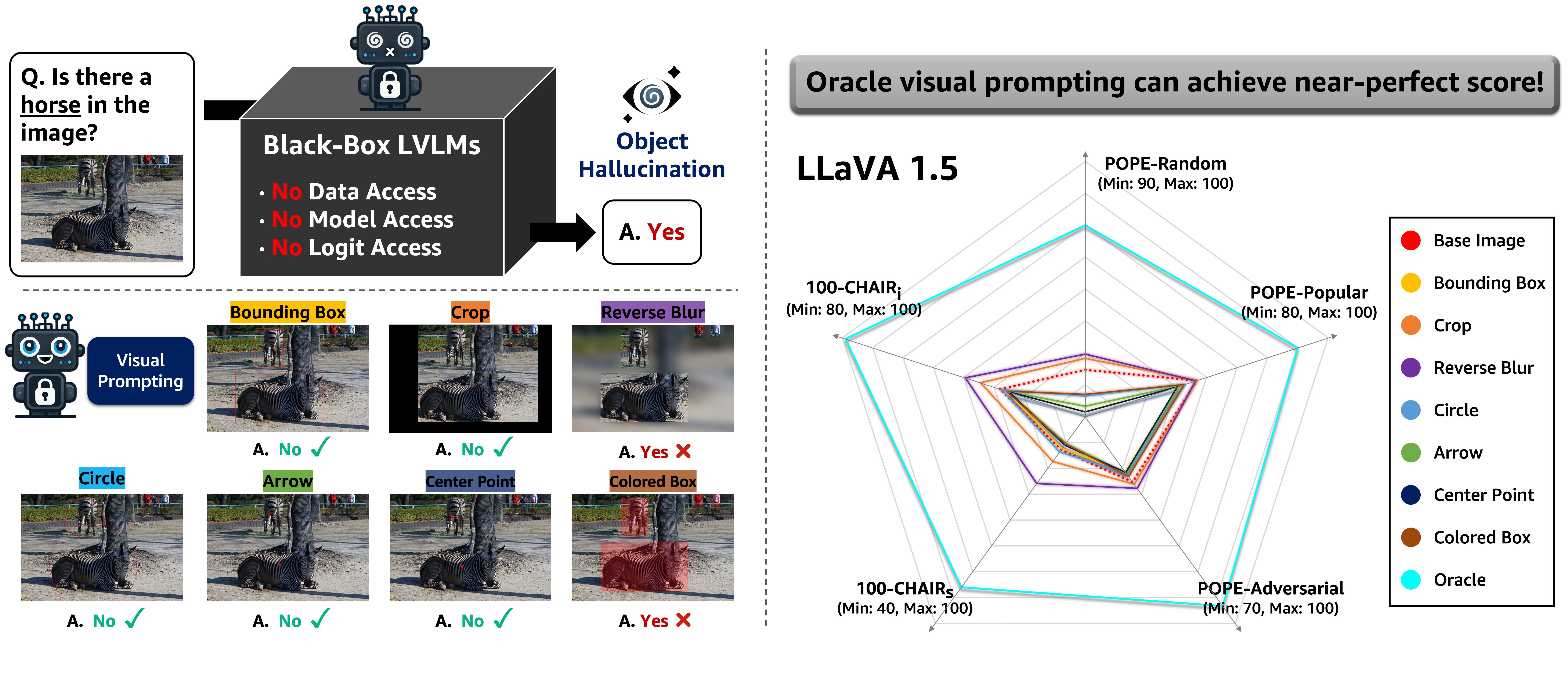}
    \vspace{-8mm}
    \caption{
        \textbf{Motivation.}
        % \textbf{(Q1) Can visual prompting mitigate object hallucinations in black-box LVLMs?}
        \textbf{(left)} An LVLM misidentifies a zebra as a horse, demonstrating object hallucination.
        % We aim to explore whether visual prompting can reduce such hallucinations.
        % We aim to explore whether visual prompting, such as overlaying bounding boxes, circles, or other visual cues on the image, can effectively reduce such hallucinations, even without direct access to model internals (\eg, data, model, or prediction logits).
        % Different visual prompts elicit varied responses, but not all are equally effective; their success depends on the specific characteristics of the image.
        % but effectiveness depends on image characteristics.
        Various VPs elicit different responses, but their effectiveness depends on the specific characteristics of the image.
        % \textbf{Oracle visual prompting can achieve near-perfect score in object hallucination benchmarks~\cite{li2023evaluating,rohrbach2018object}.}
        % Second question arises here.
        To remove randomness and solely see the impact of visual prompting, all responses are generated using greedy decoding.
        % To isolate the impact of visual prompting, we use greedy decoding for all responses.
        \textbf{(right)} While most VPs yield comparable performances, an \textit{Oracle}---which adaptively applies the best-performing VP per image---dramatically boosts results.
        % Now question arises: Can we systematically learn the optimal visual prompts without direct access to model internals (\eg, data, model, or logits)?
        % For certain metric, using a single VP for every image vs. adaptively selecting a best scoring VP for each image (Oracle).
        % \kz{how to determine the Oracle is more important here}
    }%
    \vspace{-3mm}
    \label{fig:teaser}
% \end{strip}
\end{figure*}

% \begin{figure}[t!]
%     \centering
%     \includegraphics[width=\linewidth]{figures/teaser.png}
%     \vspace{-2mm}
%     \caption{
%         \textbf{(Q1) Can visual prompting mitigate object hallucinations in black-box VLAs?}
%         In the example, a VLA erroneously identifies a zebra as a horse.
%         We aim to explore whether visual prompting, such as overlaying bounding boxes, circles, or other visual cues on the image, can effectively reduce such hallucinations, even without direct access to model internals (\eg, data, model, or prediction logits). 
%         \todo{add answers}
%     }%
%     \vspace{-3mm}
%     \label{fig:teaser}
% \end{figure}

\section{Introduction}

% Large Vision Language Models (LVLMs) have achieved significant advancements in tasks such as visual captioning, question answering, and dialog systems~\cite{tong2024cambrian,liu2023visual,bai2023qwen,chen2024far}. However, they often suffer from object hallucination, generating descriptions of objects not present in the image, undermining their reliability, particularly in critical applications like healthcare, autonomous driving, and assistive technologies~\cite{hu2024omnimedvqa,xu2024mlevlm,ma2023dolphins,shah2023lm}.

% Addressing object hallucination is crucial for deploying LVLMs in real-world scenarios.
% Existing methods to mitigate this issue involve collecting datasets~\cite{lu2024evaluation}, re-training or fine-tuning~\cite{zhao2023beyond}, modifying decoding methods~\cite{leng2023mitigating,favero2024multi}. However, these approaches require extensive computational resources and access to model internals, making them impractical for proprietary LVLMs~\cite{openai2024gpt4o,anthropic2024claude}.

% LVLMs have achieved significant advancements in tasks such as visual captioning, question answering, and dialog systems~\cite{tong2024cambrian,bai2023qwen}.
LVLMs~\cite{tong2024cambrian,bai2023qwen} demonstrate impressive capabilities but often suffer from object hallucination, where they describe objects not present in the image.
Addressing this issue is vital for real-world deployment, particularly in critical areas like healthcare and assistive technologies~\cite{hu2024omnimedvqa,xu2024mlevlm}.

Existing methods try to mitigate object hallucination by collecting datasets~\cite{lu2024evaluation}, re-training or fine-tuning~\cite{zhao2023beyond}, modifying decoding methods~\cite{leng2023mitigating,favero2024multi,woo2024ritual,woo2024don}, or using costly feedback loops~\cite{lee2023volcano}.
However, they often require access to model internals (\eg, attention, logits), making them impractical for proprietary LVLMs~\cite{openai2024gpt4o,anthropic2024claude}.

A promising yet under-explored direction is visual prompting, which overlays visual cues like bounding boxes or circles on images to guide model outputs~\cite{yao2024cpt,shtedritski2023does,yang2023fine,yang2023dawn,yang2023set}. While visual prompting has shown potential in improving visual grounding~\cite{yang2023dawn,yang2023set}, its role in reducing object hallucination remains unclear. This raises two key questions: \textbf{(Q1)} Can visual prompting mitigate object hallucination in LVLMs? \textbf{(Q2)} If so, can we systematically learn the optimal VPs?

% Surprisingly,
Our preliminary experiments show that simple object-based VPs can significantly reduce object hallucination.
Interestingly, their effectiveness varies across images and is particularly notable in an \textit{Oracle} scenario, where the best-performing VP for each image is assumed to be known.
This finding effectively answers Q1 (see~\cref{fig:teaser}) and suggests the need for a systematic method to identify the optimal VP for each image.
% \kz{we have to explain what does oracle mean when we refer to the figure 1, either in introduction or in the caption}
% \kz{which actually answers Q1}

To answer Q2, we introduce \textbf{BBVPE}, a novel framework designed to systematically identify and apply optimal VPs to reduce object hallucination in LVLMs. Our approach treats LVLMs as "black boxes", relying solely on input-output pairs without modifying the model itself. The framework has three key components: (1) a pool of predefined VPs, (2) a scoring function to evaluate the effectiveness of each prompt, and (3) a router model that dynamically selects the best prompt based on observed input-output behavior. Our method requires no access to model internals, making it applicable to both open-source and proprietary LVLMs.
% , making it particularly useful for scenarios where such access is restricted.

Our key contributions are: 
\textbf{1)} We find that Oracle VPs exist for images given an LVLM, which, when identified, can greatly reduce object hallucination.
\textbf{2)} We propose a novel framework, BBVPE, for systematically identifying these optimal VPs.
% to mitigate object hallucination in LVLMs.
\textbf{3)} In standard benchmarks like POPE and CHAIR, our approach significantly reduces object hallucination in both open-source and proprietary LVLMs.

\section{Related Work}
% \kz{we can make this section more brief to save space}
% \paragraph{Hallucinations in LVLMs.}
\noindent\textbf{Hallucinations in LVLMs.}
Efforts to address hallucination in LVLMs~\cite{dai2023instructblip,liu2023visual,liu2023improved} have focused on three primary areas:
\noindent\textit{(i) Data.}
Improving data quality is a key to reducing hallucinations~\cite{wang2023vigc}, using negative~\cite{liu2023mitigating} and counterfactual data~\cite{yu2023hallucidoctor}, as well as dataset cleansing to reduce noise and errors~\cite{yue2024less}.
\noindent\textit{(ii) Training.}
Training-based methods~\cite{jiang2023hallucination,zhai2024hallecontrol} utilize supervision from external datasets~\cite{chen2023mitigating}, reinforcement learning or preference optimization~\cite{zhao2023beyond,gunjal2024detecting} to better align model outputs with visual content.
\noindent\textit{(iii) Decoding.}
Decoding-based methods~\cite{leng2023mitigating,favero2024multi,woo2024don,woo2024ritual} refine generation by incorporating additional guidance into the output probability distribution.
% leverages external models like CLIP~\cite{radford2021learning} or DETR~\cite{carion2020end} to enhance textual accuracy.
Alternatively, post-hoc correction methods~\cite{lee2023volcano,wu2024logical,yin2023woodpecker} iteratively improve responses through self-feedback loops to identify and correct errors.
Most of these approaches assume a \textit{white-box} setting with access to model internals (\eg, data, parameters, prediction logits).
In contrast, our work addresses hallucinations in \textit{black-box} scenarios.
% and aim to reduce object hallucinations by identifying and utilizing optimal visual prompts.

% \paragraph{Automated Prompt Engineering.}
\noindent\textbf{Automated Prompt Engineering.}
% Prompt engineering enhances model performance by refining input prompts ($x$) to yield better outputs ($y^*$) without modifying model parameters ($\theta$).
Prompt engineering refines input prompts ($x$) to yield better outputs ($y^*$) without modifying model parameters ($\theta$).
% Traditionally, this has been a labor-intensive and time-consuming process.
% APE automates this prompt refinement and has been extensively explored in LLMs~\cite{shin2020autoprompt,zhou2022large,pryzant2023automatic}.
% Traditionally, this has been a labor-intensive and time-consuming process.
% APE automates this prompt refinement and has been extensively explored in LLMs~\cite{shin2020autoprompt,zhou2022large,pryzant2023automatic}.
While traditionally a manual process, APE automates this refinement and has been widely applied in LLMs~\cite{shin2020autoprompt,zhou2022large,pryzant2023automatic} to improve text prompts.
% for optimizing textual prompts.
In the vision-language domain, research has also focused on optimizing textual prompts for CLIP~\cite{liu2024language} or text-to-image diffusion models~\cite{manas2024improving,liu2024you}.
% While LLMs evolve into multimodal systems, capable of processing both textual and visual data, APE's application to visual inputs remains largely unexplored.
% To our knowledge, this is the first attempt to extend APE to the visual inputs, aiming to de-hallucinate LVLMs.
With LLMs evolve into multimodal system, capable of handling both text and visual data, APE’s application to visual inputs is still largely unexplored.
To our knowledge, this work is the first to extend APE to visual inputs, aiming to reduce hallucinations in LVLMs.

% \todo{add visual prompting}
\section{Black-Box Visual Prompt Engineering}
\begin{figure*}[t!]
    \centering
    \vspace{-3mm}
    \includegraphics[width=0.93\linewidth]{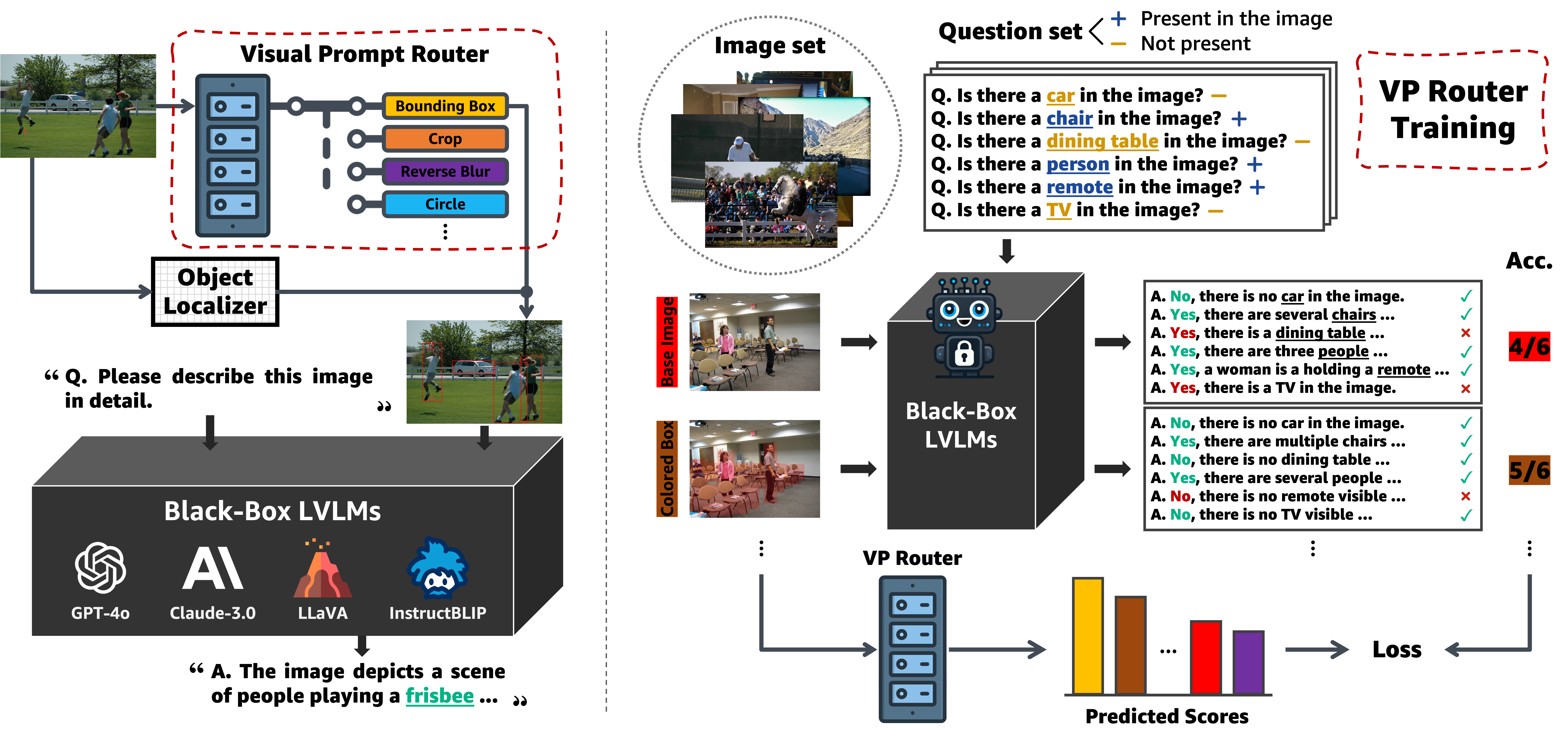}
    \vspace{-4mm}
    \caption{
        \textbf{Overview.}
        \textbf{(left)} BBVPE utilizes a VP router and object localizer to mitigate object hallucinations in LVLMs.
        VP router dynamically selects the optimal VP for a given image.
        \textbf{(right)} During its training phase, a set of images with various VPs and a series of object-related questions are posed to the LVLMs.
        The question set includes both objects that are present and not present in the image.
        LVLM responses are then evaluated based on accuracy.
        The VP router predicts scores for each VP, optimizing the selection process to identify the most effective prompt for a given image.
    }%
    \vspace{-4mm}
    \label{fig:overview}
\end{figure*}

% \kz{we should demonstrate our oracle experiments somewhere and highlight the novelty compared to previous visual prompting works}
% \noindent\textbf{Overview.}
Applying prompt engineering to the visual domain is challenging due to the vast combinatorial complexity of image space.
Also, direct optimization over pixel values risks distorting the semantic content of the images.
To circumvent this, we use a discrete selection approach, choosing from a predefined VPs that enhance images without altering their original meaning.
A lightweight router model selects the most suitable VP, which is then applied before input to LVLMs, reducing hallucinations.
Our black-box approach mitigates hallucinations without accessing internal LVLM values (\eg, attention, logits), making it compatible with proprietary models.
An overview is shown in~\cref{fig:overview}.
% based on the LVLM’s preference.
% This automated, model-agnostic approach requires no human intervention and works with both open-source and proprietary LVLMs.
% \footnote{For more details about our framework, see~\cref{sec:method_details}.}
% , including LLaVA 1.5~\cite{liu2023improved}, InstructBLIP~\cite{dai2023instructblip}, GPT-4o~\cite{openai2024gpt4o}, and Claude-3.0-Sonnet~\cite{anthropic2024claude}.

\noindent\textbf{Oracle.}
The Oracle represents an ideal scenario where the optimal VP for each image is known during evaluation, setting an upper bound on performance (see~\cref{fig:teaser} right).
It is equivalent to adaptively selecting the VP with minimal hallucination per image.
Our goal is to train the router model to approximate this behavior.

%%%%%%%%%%%%%%%%%%%%%%%%%%
% kang's edition
% \vspace{1mm}
\noindent\textbf{Object localization.}
To identify relevant objects within an image \( I \), we first utilize an object localization model \( \mathcal{L} \).
The model detects and outputs a set of object coordinates \( O = \{ o_1, o_2, \ldots, o_m \} \).
% We utilize an object localization model \( \mathcal{L} \) to obtain position information of objects within the image \( I \).
% The model \( \mathcal{L} \) outputs a set of object positions \( O = \{ o_1, o_2, \ldots, o_m \} \), where each \( o_j \) represents the coordinates of an object in the image.

% \vspace{1mm}
\noindent\textbf{Visual prompt pool.}
We define a pool of candidate VPs \( P = \{ p_1, p_2, \ldots, p_n \} \), which includes visual markers like circles and arrows.
Each VP \( p_i \in P \) modifies the image \( I \) by highlighting localized objects \( O \), producing \( I_{p_i} \).
The image-text pair \( (I_{p_i}, T) \), where \( T \) is a textual prompt, is then fed into the LVLM \( \mathcal{M} \) to produce a response.
% The VLA \( M \) takes as input both the modified image \( I_{p_i} \) and a textual prompt \( T \). 
% The textual prompt \( T \) varies depending on the scoring function \( S \) used:
    % \begin{itemize}
%     \item For the CHAIR metric, \( T \) is set to \textit{``Please describe the image in detail.''}
%     \item For the POPE metric, \( T \) is set to \textit{``Is there an \{object\} in the image?''}
% \end{itemize}
% To evaluate object hallucination, we employ a score function \( S \), which quantifies the level of hallucination in the model's response.

\noindent\textbf{Quantifying object hallucination.}
To evaluate a model’s robustness to object hallucination, we define a scoring function \( S \) that measures response accuracy regarding object presence:
\begin{equation}
S = \frac{|\text{correct responses}|}{|\text{total presence questions}|}
\end{equation}
% To quantify object hallucination, we introduce a scoring function \( S \), which evaluates the robustness of the model's response.
% In practice, we ask both positive and negative presence-related questions about objects and measure the accuracy of responses.
% Simply, the score \( S \) is defined as follows:
% To assess object hallucination, we use a scoring function \( S \) that measures the model's robustness to hallucination by asking object presence questions and evaluating response accuracy. 

\noindent\textbf{Dataset construction.}
For a given image \( I \), the optimal VP \( p^* \) is chosen to maximize \( S \):
\begin{equation}
p^* = \arg\max_{p_i \in P} S\left( \mathcal{M}\left( I_{p_i}, T \right) \right)
\end{equation}
To ensure uniqueness, cases where multiple VPs achieve the highest score are excluded.
This results in a training dataset \( D_{\text{train}} \) that maps images to unique optimal prompts, including the option of not applying any VP:
\begin{equation}
D_{\text{train}} = \left\{ \left( I_j, p_j^* \right) \mid \text{unique } p_j^* \right\}
\end{equation}
% In cases where multiple VPs achieve the best score, we exclude those examples to ensure that each image in our training dataset \( D_{\text{train}} \) is associated with a unique optimal VP (including the option of not applying any prompt).
% 
% If multiple VPs achieve the best score, we exclude those cases to ensure that each image in our training dataset \( D_{\text{train}} \) is associated with a unique optimal VP (including the option of no VPs).
% This results in a mutually exclusive subset of data:
% 

\noindent\textbf{Training a router model.}
The router model \( \mathcal{R}_{\theta} \) is trained on \( D_{\text{train}} \) to predict the optimal VP \( p^* \) for a given image \( I \).
It assigns a score \( \hat{s}_{p_i} \) to each VP:
\begin{equation}
\hat{s}_{p_i} = \mathcal{R}_{\theta}\left( I, p_i \right)
\end{equation}
These scores are converted into probabilities via softmax:
\begin{equation}
\hat{P}(p_i \mid I) = \frac{\exp(\hat{s}_{p_i})}{\sum_{p_j \in P} \exp(\hat{s}_{p_j})}
\end{equation}
The router model is trained using cross-entropy loss between the predicted probability distribution \( \hat{P}(p_i \mid I) \) and the one-hot encoded ground-truth optimal VP \( p^* \):
\begin{equation}
\vspace{-1mm}
\mathcal{L} = - \sum_{p_i \in P} \mathbbm{1}_{p_i = p^*} \log \hat{P}(p_i \mid I)
\vspace{-1mm}
\end{equation}
The trained router model enables efficient VP selection without directly querying the LVLM.
% \( \mathcal{R} \) takes only the image \( I \) as input and does not utilize the textual prompt \( T \).
% \begin{equation}
% \hat{s}_{p_i} = \mathcal{R}\left( I, p_i \right)
% \end{equation}
% We formulate this as a regression problem, where the model predicts a score \( \hat{s}_{p_i} \) for each possible VP \( p_i \), and the prompt with the highest predicted effectiveness is selected:

% \noindent\textbf{Applying the predicted VP.}
\noindent\textbf{LVLM inference.}
At inference, the trained router model \( \mathcal{R}_{\theta} \) predicts the optimal VP \( \hat{p} \):
\begin{equation}
\vspace{-1mm}
\hat{p} = \arg\max_{p_i \in P} \hat{s}_{p_i}
\vspace{-1mm}
\end{equation}
Applying \( \hat{p} \) to the localized objects \( O \) in \( I \) produces \( I_{\hat{p}} \), which, along with the textual prompt \( T \), is fed into LVLM \( \mathcal{M} \) to obtain a response with reduced object hallucination.
\begin{table*}[t!]
    \centering
    \small
    \setlength{\tabcolsep}{2pt} % base value: 6pt
    \renewcommand{\arraystretch}{1.05}
    \vspace{-2mm}
    \scalebox{0.72}{
    \begin{tabular}{y{25}y{50}x{30}x{30}x{30}x{30}x{30}x{30}x{30}x{30}x{30}x{30}x{30}x{30}x{30}x{30}x{30}x{30}}
    \toprule
     &  & \multicolumn{8}{c}{\textbf{Open-source LVLMs}} & \multicolumn{8}{c}{\textbf{Proprietary LVLMs}} \\
    \arrayrulecolor{gray} \cmidrule(lr){3-10} \cmidrule(lr){11-18}
    \multirow{2}{*}{\textbf{Setup}} & \multirow{2}{*}{\textbf{Methods}} & \multicolumn{4}{c}{\textbf{LLaVA 1.5}} & \multicolumn{4}{c}{\textbf{InstructBLIP}} & \multicolumn{4}{c}{\textbf{GPT-4o}} & \multicolumn{4}{c}{\textbf{Claude-3.0-Sonnet}} \\
    \arrayrulecolor{gray} \cmidrule(lr){3-6} \cmidrule(lr){7-10} \cmidrule(lr){11-14} \cmidrule(lr){15-18}
     &  & Acc.{\up} & Prec.{\up} & Rec.{\up} & F1{\up} & Acc.{\up} & Prec.{\up} & Rec.{\up} & F1{\up} & Acc.{\up} & Prec.{\up} & Rec.{\up} & F1{\up} & Acc.{\up} & Prec.{\up} & Rec.{\up} & F1{\up} \\
    \midrule
    \multirow{6}{*}{~~~\rot{Random}} 
     & \textit{baseline} & $89.60$ & $88.77$ & $90.67$ & $89.71$ & $90.23$ & \colorbox{Best}{$92.95$} & $87.07$ & $89.91$ & $87.33$ & $97.95$ & $76.27$ & $85.76$ & $79.93$ & \colorbox{Best}{$98.18$} & $61.00$ & $75.25$ \\
     & \textit{random VP} & $89.46$ & $89.07$ & $89.95$ & $89.51$ & $89.75$ & $91.76$ & $87.35$ & $89.50$ & $87.02$ & $96.63$ & $76.75$ & $85.53$ & $78.91$ & $97.74$ & $59.18$ & $73.71$ \\
     & \textit{best VP$^{\dagger}$} & $90.40$ & $90.67$ & $90.07$ & $90.37$ & $89.97$ & $91.89$ & $87.67$ & $89.73$ & $88.07$ & $98.47$ & $77.33$ & $86.63$ & $80.10$ & $97.78$ & $61.60$ & $75.58$ \\
     & \textbf{BBVPE} & \colorbox{Best}{$91.37$} & \colorbox{Best}{$91.97$} & \colorbox{Best}{$91.40$} & \colorbox{Best}{$91.42$} & \colorbox{Best}{$91.50$} & $90.47$ & \colorbox{Best}{$91.44$} & \colorbox{Best}{$90.95$} & \colorbox{Best}{$88.83$} & \colorbox{Best}{$98.71$} & \colorbox{Best}{$78.26$} & \colorbox{Best}{$87.31$} & \colorbox{Best}{$80.84$} & $97.43$ & \colorbox{Best}{$63.49$} & \colorbox{Best}{$76.88$} \\
    \arrayrulecolor{gray!50}\cmidrule(lr){2-18}
     & \textit{Oracle} & $93.99$ & $95.13$ & $94.69$ & $93.94$ & $94.04$ & $97.16$ & $92.46$ & $93.44$ & $93.50$ & $99.47$ & $87.48$ & $93.09$ & $85.87$ & $99.27$ & $72.27$ & $83.64$ \\
    \midrule
    \multirow{6}{*}{~~~\rot{Popular}} 
     & \textit{baseline} & $86.20$ & $83.23$ & \colorbox{Best}{$90.67$} & $86.79$ & $83.43$ & $81.17$ & $87.07$ & $84.01$ & $86.03$ & $94.56$ & $76.47$ & $84.56$ & $78.43$ & $93.56$ & $61.07$ & $73.90$ \\
     & \textit{random VP} & $86.20$ & $83.68$ & $89.96$ & $86.70$ & $83.12$ & $80.54$ & $87.35$ & $83.80$ & $85.26$ & $92.38$ & $76.91$ & $83.92$ & $77.48$ & $93.24$ & $59.24$ & $72.44$ \\
     & \textit{best VP$^{\dagger}$} & $86.70$ & $84.38$ & $90.07$ & $87.13$ & $84.13$ & $81.88$ & $87.67$ & $84.67$ & $86.37$ & $94.31$ & $77.40$ & $85.02$ & $78.70$ & $93.90$ & $61.60$ & $74.40$ \\
     & \textbf{BBVPE} & \colorbox{Best}{$87.23$} & \colorbox{Best}{$85.97$} & $90.20$ & \colorbox{Best}{$88.03$} & \colorbox{Best}{$84.57$} & \colorbox{Best}{$82.41$} & \colorbox{Best}{$88.71$} & \colorbox{Best}{$85.44$} & \colorbox{Best}{$87.33$} & \colorbox{Best}{$95.31$} & \colorbox{Best}{$79.22$} & \colorbox{Best}{$86.52$} & \colorbox{Best}{$79.67$} & \colorbox{Best}{$94.90$} & \colorbox{Best}{$62.42$} & \colorbox{Best}{$75.30$} \\
    \arrayrulecolor{gray!50}\cmidrule(lr){2-18}
     & \textit{Oracle} & $91.97$ & $92.81$ & $94.69$ & $92.38$ & $88.52$ & $89.65$ & $92.46$ & $89.06$ & $92.57$ & $98.04$ & $86.87$ & $92.12$ & $84.87$ & $96.78$ & $72.13$ & $82.66$ \\
    \midrule
    \multirow{6}{*}{~~~\rot{Adversarial}}
     & \textit{baseline} & $79.73$ & $74.40$ & $90.67$ & $81.73$ & $80.73$ & $77.28$ & $87.07$ & $81.88$ & $85.50$ & \colorbox{Best}{$93.33$} & $76.47$ & $84.06$ & $77.13$ & \colorbox{Best}{$89.82$} & $61.20$ & \colorbox{Best}{$72.80$} \\
     & \textit{random VP} & $79.56$ & $74.48$ & $89.95$ & $81.49$ & $79.87$ & $75.99$ & $87.35$ & $81.27$ & $84.49$ & $90.76$ & $76.85$ & $83.20$ & $75.90$ & $88.83$ & $59.25$ & $71.07$ \\
     & \textit{best VP$^{\dagger}$} & $80.30$ & $75.35$ & $90.07$ & $82.05$ & $80.20$ & $76.28$ & $87.67$ & $81.58$ & $85.73$ & $93.07$ & $77.00$ & $84.28$ & $76.90$ & $88.76$ & \colorbox{Best}{$61.60$} & $72.73$ \\
     & \textbf{BBVPE} & \colorbox{Best}{$81.33$} & \colorbox{Best}{$75.84$} & \colorbox{Best}{$91.77$} & \colorbox{Best}{$83.05$} & \colorbox{Best}{$81.23$} & \colorbox{Best}{$77.33$} & \colorbox{Best}{$88.49$} & \colorbox{Best}{$82.53$} & \colorbox{Best}{$86.00$} & $92.19$ & \colorbox{Best}{$78.67$} & \colorbox{Best}{$84.89$} & \colorbox{Best}{$78.00$} & $88.89$ & $61.54$ & $72.73$ \\
    \arrayrulecolor{gray!50}\cmidrule(lr){2-18}
     & \textit{Oracle} & $85.62$ & $84.23$ & $94.69$ & $87.25$ & $85.72$ & $85.98$ & $92.46$ & $86.80$ & $91.90$ & $96.94$ & $86.53$ & $91.44$ & $83.53$ & $94.36$ & $71.33$ & $81.25$ \\
     \bottomrule
    \end{tabular}
    }
    \vspace{\abovetabcapmargin}
    \vspace{-1mm}
    \caption{
        \textbf{Results on POPE benchmark.}
        Our approach consistently outperforms baselines; yet, there is still a large gap compared to \textit{Oracle}.
        ${\dagger}$ Best VPs are: \texttt{`reverse blur'} for LLaVA and InstructBLIP, \texttt{`crop'} for GPT-4o and Claude-3.0-Sonnet.
    }
    \vspace{\belowtabcapmargin}
    \label{tab:pope_combined}
\end{table*}

\section{Experiments}
In all tables, \textit{baseline} refers to not using visual prompting.
% As baselines, we compare our approach with selecting \textit{random} visual prompts, using the \textit{best singular} VP that achieves the best overall performance for certain model, and an Oracle as the ideal performance scenario.
We compare our approach against three baselines: (1) selecting \textit{random VP} for each image, (2) consistently using a fixed \textit{best VP} that delivers the highest overall performance for the model, and (3) an \textit{Oracle} that adaptively selects the optimal VP per image.
% \kz{better to explain in a bit more detail, even for me, I could not immediately get the idea}
Responses are generated via greedy decoding to eliminate randomness.\footnote{Implementation details are in~\cref{sec:implementation_details}.}

% To ensure there is no randomness, all experiments used standard greedy decoding, which directly samples the argmax token from the probability distribution while generating response.

% \kz{we should include the oracle experiments}

\vspace{1mm}\noindent\textbf{Evaluation setup.}
We evaluate using POPE~\cite{li2023evaluating} and CHAIR~\cite{rohrbach2018object} on the COCO~\cite{lin2014microsoft} val split.
POPE assesses hallucination by asking binary Yes/No questions like "Is there a [object] in the image?" across various prompt setups (Random, Popular, and Adversarial).
CHAIR measures the ratio of hallucinated objects in image descriptions, with two variants: CH$_S$ (per sentence) and CH$_I$ (per object), where lower scores indicate fewer hallucinations.
Additionally, we use GPT-4o~\cite{openai2024gpt4o} for a more comprehensive evaluation.\footnote{More details about evaluation setup are in~\cref{sec:evaluation_details}.}
% \kz{should refer reviewers to appendix for metric details}

\vspace{1mm}\noindent\textbf{Model instantiation.}
While our framework is generic, we instantiate the components as follows:
\begin{itemize}[leftmargin=*, labelsep=0.3em]
    \vspace{-2.5mm}
    \item \textbf{Object Localizer \( \mathcal{L} \)}: SAM 2~\cite{ravi2024sam}.
    \vspace{-2.5mm}
    \item \textbf{VP Router \( \mathcal{R}_{\theta} \)}: Frozen CLIP vision encoder~\cite{radford2021learning} with a trainable MLP.
    \vspace{-7.5mm}
    \item \textbf{LVLMs \( \mathcal{M} \)}: We use two open-source models (LLaVA-1.5, InstructBLIP) and two proprietary models (GPT-4o, Claude-3.0-Sonnet).
    % \vspace{-2mm}\item \textbf{Score Function \( S \)}: POPE~\cite{li2023evaluating}.
\vspace{-2.5mm}
\end{itemize}
During router training, all other model components are kept frozen.

\begin{table*}[t!]
    \begin{minipage}[t!]{0.59\linewidth}
        \begin{center}
        \begin{small}
        \setlength{\tabcolsep}{1pt} % base value: 6pt
        \scalebox{0.81}{
        \begin{tabular}{lx{33}x{33}x{33}x{33}x{33}x{33}x{33}x{33}}
            \toprule
             & \multicolumn{4}{c}{\textbf{Open-source LMMs}} & \multicolumn{4}{c}{\textbf{Proprietary LMMs}} \\
            \cmidrule(lr){2-5} \cmidrule(lr){6-9}
            \multirow{2}{*}{\textbf{Methods}} & \multicolumn{2}{c}{\textbf{LLaVA 1.5}} & \multicolumn{2}{c}{\textbf{InstructBLIP}} & \multicolumn{2}{c}{\textbf{GPT-4o}} & \multicolumn{2}{c}{\textbf{Claude-3.0}} \\
             \cmidrule(lr){2-3} \cmidrule(lr){4-5} \cmidrule(lr){6-7} \cmidrule(lr){8-9}
             & CH$_S${\down} & CH$_I${\down} & CH$_S${\down} & CH$_I${\down} & CH$_S${\down} & CH$_I${\down} & CH$_S${\down} & CH$_I${\down} \\
            \midrule
            \textit{baseline}  & $62.8$ & $18.1$ & $53.6$ & $14.7$ & $44.9$ & $8.0$ & $38.5$ & $12.1$ \\
            \textit{random VP} & $61.7$ & $18.4$ & $53.7$ & $15.8$ & $45.2$ & $8.0$ & $39.0$ & $13.9$ \\
            \textit{best VP$^{\dagger}$} & $56.3$ & $17.0$ & $48.5$ & $14.4$ & $36.5$ & $5.9$ & $33.9$ & $11.4$ \\
            \textbf{BBVPE} & \colorbox{Best}{$46.3$} & \colorbox{Best}{$14.9$} & \colorbox{Best}{$41.5$} & \colorbox{Best}{$12.5$} & \colorbox{Best}{$32.0$} & \colorbox{Best}{$4.9$} & \colorbox{Best}{$31.7$} & \colorbox{Best}{$10.7$} \\
            \arrayrulecolor{gray!50}\cmidrule(lr){1-9}
            \textit{Oracle} & $27.7$ & $6.4$ & $18.5$ & $3.8$ & $8.4$ & $1.3$ & $7.4$ & $2.0$ \\
            \bottomrule
        \end{tabular}
        }
        \end{small}
        \end{center}
        \vspace{\abovetabcapmargin}
        \vspace{-2mm}
        \captionof{table}{
            \textbf{Results on CHAIR benchmark.}
            Black-Box VPE significantly reduces hallucinations in image descriptions.
            ${\dagger}$ Best VPs are: \texttt{`center point'} for LLaVA and InstructBLIP, \texttt{`reverse blur'} for GPT-4o, and \texttt{`arrow'} for Claude-3.0-Sonnet.
        }
        \vspace{\belowtabcapmargin}
        \label{tab:chair_combined}
    \end{minipage}
    \hfill
    \begin{minipage}[t!]{0.39\linewidth}
        \begin{center}
        \begin{small}
        \setlength{\tabcolsep}{1pt} % base value: 6pt
        \scalebox{0.8}{
        \begin{tabular}{lx{26}x{26}x{26}x{26}x{26}x{26}}
            \toprule
            \multirow{2}{*}{\textbf{Methods}} & \multicolumn{6}{c}{\textbf{LLaVA 1.5}} \\
            \arrayrulecolor{gray} \cmidrule(lr){2-7}
             & Acc{\up} & Det{\up} & Com{\up} & Rel{\up} & Rob{\up} & Total{\up} \\
            \midrule
            \textit{baseline} & $7.08$ & $6.63$ & $6.67$ & $7.35$ & $7.51$ & $35.24$ \\
            \textit{random VP} & $6.38$ & $6.21$ & $6.25$ & $6.85$ & $6.84$ & $32.52$ \\
            \textit{best VP$^{\dagger}$} & $6.53$ & $6.30$ & $6.34$ & $6.92$ & $6.92$ & $33.00$ \\
            \textbf{BBVPE} & \colorbox{Best}{7.24} & \colorbox{Best}{$6.86$} & \colorbox{Best}{$6.95$} & \colorbox{Best}{$7.63$} & \colorbox{Best}{$7.70$} & \colorbox{Best}{$36.38$} \\
            \arrayrulecolor{gray!50}\cmidrule(lr){1-7}
            \textit{Oracle} & $7.59$ & $7.27$ & $7.30$ & $8.03$ & $8.10$ & $38.29$ \\
            \bottomrule
        \end{tabular}
        }
        \end{small}
        \end{center} 
        \vspace{\abovetabcapmargin}
        \vspace{-1mm}
        \captionof{table}{
            \textbf{Comprehensive image description evaluation by GPT-4o.}
            LLaVA is assessed based on 5 criteria: \underline{Acc}uracy, \underline{Det}ail, \underline{Com}prehensiveness, \underline{Rel}evance, and \underline{Rob}ustness.
            ${\dagger}$ The best VP is \texttt{`center point'}.
        }
        \vspace{\belowtabcapmargin}
        \label{tab:gpt_evaluation}
    \end{minipage}
    \vspace{-2mm}
\end{table*}

\subsection{Evaluation Results}

\noindent\textbf{POPE benchmark.}
\cref{tab:pope_combined} shows BBVPE consistently outperforms baselines across most metrics, prompt setups, and LVLMs.
While \textit{random VP} may not improve results over \textit{baseline} (No VP applied), \textit{best VP} generally performs better.
BBVPE further enhances performance by properly routing the optimal VP for each image, though a gap remains to \textit{Oracle}, suggesting room for improvement.

\vspace{1mm}\noindent\textbf{CHAIR benchmark.}
As shown in~\cref{tab:chair_combined}, BBVPE significantly reduces object hallucinations in image descriptions at both instance (CH$_I$) and sentence (CH$_S$) levels across all LVLMs, though still below \textit{Oracle} performance.
While \textit{random VP} often underperforms \textit{baseline}, \textit{best VP} consistently improves results, with BBVPE further enhancing performance.

\vspace{1mm}\noindent\textbf{GPT-4o evaluation.}
% We use GPT-4o for a more comprehensive evaluation of object hallucinations.
\cref{tab:gpt_evaluation} shows GPT-4o's evaluation of image descriptions from LLaVA 1.5, scored from 0 to 10.
GPT-4o receives the image and the generated descriptions, scoring each based on 5 criteria.\footnote{Details on GPT-4o instruction are in~\cref{sec:gpt4o_evaluation}.}
% Here, naively applying visual prompting rather harms the performance. 
% However, BBVPE can achieve the best scores by properly routing optimal VPs.
While naive visual prompting (\textit{random VP}, \textit{best VP}) degrade performance, BBVPE effectively improves scores.
Notably, applying a fixed \textit{best VP} to all images performs even worse than using no VP (\textit{baseline}), but BBVPE outperforms both by optimally selecting VPs per image.

\subsection{Key Observations}

\noindent(1) Different LVLMs favor different VPs.
For example, \texttt{`reverse blur'} and \texttt{`crop'} generally work well for LLaVA 1.5 (\cref{fig:teaser} (Right)).
% \kz{cannot determine which are which in right sub-figure}

\noindent(2) Surprisingly, proprietary LVLMs underperform compared to open-source LVLMs on POPE in terms of Accuracy and F1 score (\cref{tab:pope_combined}).
Proprietary LVLMs are cautious to say "yes"---indicated by high precision but low recall.
It suggests a conservative response strategy, likely due to policy restrictions aimed at minimizing false positives.
% \kz{can we merge 2 and 3}
% intended to respond sensitively to false positive information.

% \noindent(3) Proprietary LVLMs are more robust against adversarial prompt setup in POPE, while open-source LVLMs are more susceptible (see~\cref{tab:pope_combined}).\kz{IMO, better to remove, too detailed}

\noindent(3) No single VP achieves optimal results across all LVLMs and metrics; the best VP varies by model and metric. (\cref{tab:pope_combined,tab:chair_combined,tab:gpt_evaluation})
% \kz{no comparison between VPs from the tables}
% \todo{best singular VP}

\noindent(4) Learning an effective routing of VPs can significantly reduce hallucinations (\cref{tab:pope_combined,tab:chair_combined,tab:gpt_evaluation}).

% \input{figures/04_attention}
% \paragraph{Attention visualization.}
% \cref{fig:attention} suggest that visual prompting can direct the LVLM’s attention to relevant objects, thereby reducing hallucinations.

\subsection{Analysis}
\noindent\textbf{Computational cost.}
We analyze the latency and computational overhead (TFLOPs) of recent methods for object hallucination mitigation in~\cref{tab:computational_cost}.
% VCD~\cite{liu2023mitigating} applies diffusion noise to the original image and requires two forward passes for contrastive decoding.
% M3ID~\cite{favero2024multi} requires two forward passes for contrastive decoding.
% RITUAL~\cite{woo2024ritual} applies random image transformations and require two forward passes.
% AvisC~\cite{woo2024don} detects blind tokens via layer selection and requires two forward passes for contrastive decoding.
% OPERA~\cite{huang2023opera} uses beam search with frequent rollbacks to anchor tokens.
% VOLCANO~\cite{lee2023volcano} needs critique-revise-decide steps for self-feedback, adding at least three forward passes.
% BBVPE runs object localizer (\eg, SAM2) and VP router (\eg, CLIP+MLP).
VCD~\cite{liu2023mitigating}, M3ID~\cite{favero2024multi}, RITUAL~\cite{woo2024ritual}, and AvisC~\cite{woo2024don} require two forward passes, while OPERA~\cite{huang2023opera} uses beam search with rollbacks, and VOLCANO~\cite{lee2023volcano} performs critique-revise-decide steps, needing three forward passes.
BBVPE introduces some additional latency due to the use of an object localizer (\eg, SAM2) and VP router (\eg, CLIP+MLP). 
However, it is significantly more efficient than other methods. 
Unlike others relying on model internals (\eg, weights, logits), BBVPE operates in a black-box manner, making it applicable to both open-source and proprietary models.
% This flexibility represents a substantial advantage in real-world scenarios.
\begin{table}[t!]
\centering
\small
\setlength{\tabcolsep}{2pt} % base value: 6pt
\renewcommand{\arraystretch}{1.05}
\scalebox{0.85}{
\begin{tabular}{y{130}x{50}x{40}x{20}}
\toprule
\multirow{2}{*}{\textbf{Methods}} & \textbf{Latency (ms/token)} & \multirow{2}{*}{\textbf{TFLOPs}} & \multirow{2}{*}{\textbf{\faBox}} \\
\arrayrulecolor{gray!50}\cmidrule(lr){1-4} 
\cellcolor{gray!10} Baseline (LLaVA-1.5) & \cellcolor{gray!10} $43.664$ & \cellcolor{gray!10} $9.726$ & \cellcolor{gray!10} - \\ 
\quad + VCD~\cite{liu2023mitigating} & $111.392$ & $19.452$ & \xmark \\ 
\quad + M3ID~\cite{favero2024multi} & $84.49$ & $19.452$ & \xmark \\ 
\quad + RITUAL~\cite{woo2024ritual} & $88.582$ & $19.452$ & \xmark \\ 
\quad + AvisC~\cite{woo2024don} & $88.127$ & $19.452$ & \xmark \\ 
\quad + OPERA~\cite{huang2023opera} & $159.615$ & $48.628$ & \xmark \\ 
\quad + VOLCANO~\cite{lee2023volcano} & $202.122$ & $42.794$ & \xmark \\ 
\quad \textbf{+ BBVPE (Ours)} & $65.505$ & $16.968$ & \cmark \\
\bottomrule
\end{tabular}
}
\vspace{\abovetabcapmargin}
\caption{
\textbf{Comparison of methods on latency, TFLOPs, and applicability to black-box LVLMs (\textbf{\faBox}).} All runs use a single NVIDIA A100 40GB GPU.
}
\vspace{\belowtabcapmargin}
\label{tab:computational_cost}
\end{table}

\vspace{1mm}\noindent\textbf{Cross-dataset evaluation on POPE-GQA benchmark.}
% While our primary evaluation is on the COCO dataset, we conducted additional experiments on the GQA dataset to assess cross-dataset generalization.
\Cref{tab:cross_dataset} shows the results on POPE benchmark using GQA dataset.
The overall performance trends are similar to the LLaVA-1.5 results in~\cref{tab:pope_combined}.
Notably, the VP router trained on COCO performs effective VP selection even on unseen datasets like GQA, outperforming a fixed best VP and achieving results comparable to a VP router trained and tested on GQA.
This demonstrates BBVPE's potential for cross-dataset generalization.
% Our router model generalizes well to both unseen tasks and datasets, demonstrating its robustness and adaptability.
% \input{figures/04_attention}

% \vspace{1mm}\noindent\textbf{Attention visualization.}
% When properly applied, visual prompting can direct the LVLM's attention towards the relevant objects.
% This guides the model to carefully look into the prompted regions, thereby reducing the likelihood of incorrect identifications.
% For example, as shown in~\cref{fig:attention}, LLaVA-1.5 initially provides an incorrect answer, "Yes".
% However, after applying Center Point VP, the model attends to previously overlooked regions, leading to a shift in the top-10 scores and ultimately correcting the answer to "No".
% In some cases, visual prompting directs the LVLM’s attention on relevant object areas, as shown in \cref{fig:attention}, which in turn reduces hallucinations.
% LLaVA 1.5 corrects an initial incorrect answer ("Yes") to the question, “Is there a TV in the image?”, changing it to the correct response, “No”.
% Visual attention maps and top-10 scores illustrate how prompting shifts focus, improving accuracy in the model’s predictions.
% \todo{statistics across images?}
\begin{table}[t!]
\centering
\small
\setlength{\tabcolsep}{2pt} % base value: 6pt
\renewcommand{\arraystretch}{1.05}
\scalebox{0.8}{
\begin{tabular}{y{90}x{25}x{25}x{25}x{25}x{25}x{25}}
\toprule
\multirow{2}{*}{\parbox{90pt}{\textbf{Methods} \\  \textbf{(Model: LLaVA-1.5)}}} & 
\multicolumn{2}{c}{Random} & \multicolumn{2}{c}{Popular} & \multicolumn{2}{c}{Adversarial}\\
\arrayrulecolor{gray!50}\cmidrule(lr){2-3} \cmidrule(lr){4-5} \cmidrule(lr){6-7} 
 & Acc. & F1 & Acc. & F1 & Acc. & F1 \\
\arrayrulecolor{gray!50}\cmidrule(lr){1-7} 
\textit{baseline} & 81.23 & 83.16 & 72.43 & 77.31 & 69.07 & 75.37 \\ 
\textit{random VP} & 80.97 & 82.95 & 72.07 & 77.00 & 68.70 & 74.94 \\ 
\textit{best VP} (reverse blur) & 82.10 & 83.99 & 73.27 & 78.02 & 69.43 & 75.43 \\ 
\arrayrulecolor{gray!50}\cmidrule(lr){1-7} 
\multicolumn{7}{l}{\textbf{BBVPE} (train dataset $\rightarrow$ test dataset)} \\ 
\quad GQA $\rightarrow$ GQA & \textbf{83.47} & \textbf{84.89} & \textbf{74.37} & \textbf{78.56} & \textbf{71.73} & \textbf{76.87} \\ 
\quad COCO $\rightarrow$ GQA & \textbf{82.73} & \textbf{84.17} & \textbf{73.83} & \textbf{78.28} & \textbf{70.30} & \textbf{75.90} \\ 
\arrayrulecolor{gray!50}\cmidrule(lr){1-7} 
Oracle & 92.93 & 93.05 & 82.27 & 84.00 & 76.87 & 80.21 \\
\bottomrule
\end{tabular}
}
\vspace{\abovetabcapmargin}
\caption{
\textbf{Results on POPE benchmark using GQA dataset~\cite{hudson2019gqa}.} Here, we also compare with cross-dataset evaluation setup (COCO $\rightarrow$ GQA).
}
\vspace{\belowtabcapmargin}
\label{tab:cross_dataset}
\end{table}

\begin{figure}[t!]
    \centering
    \includegraphics[width=\linewidth]{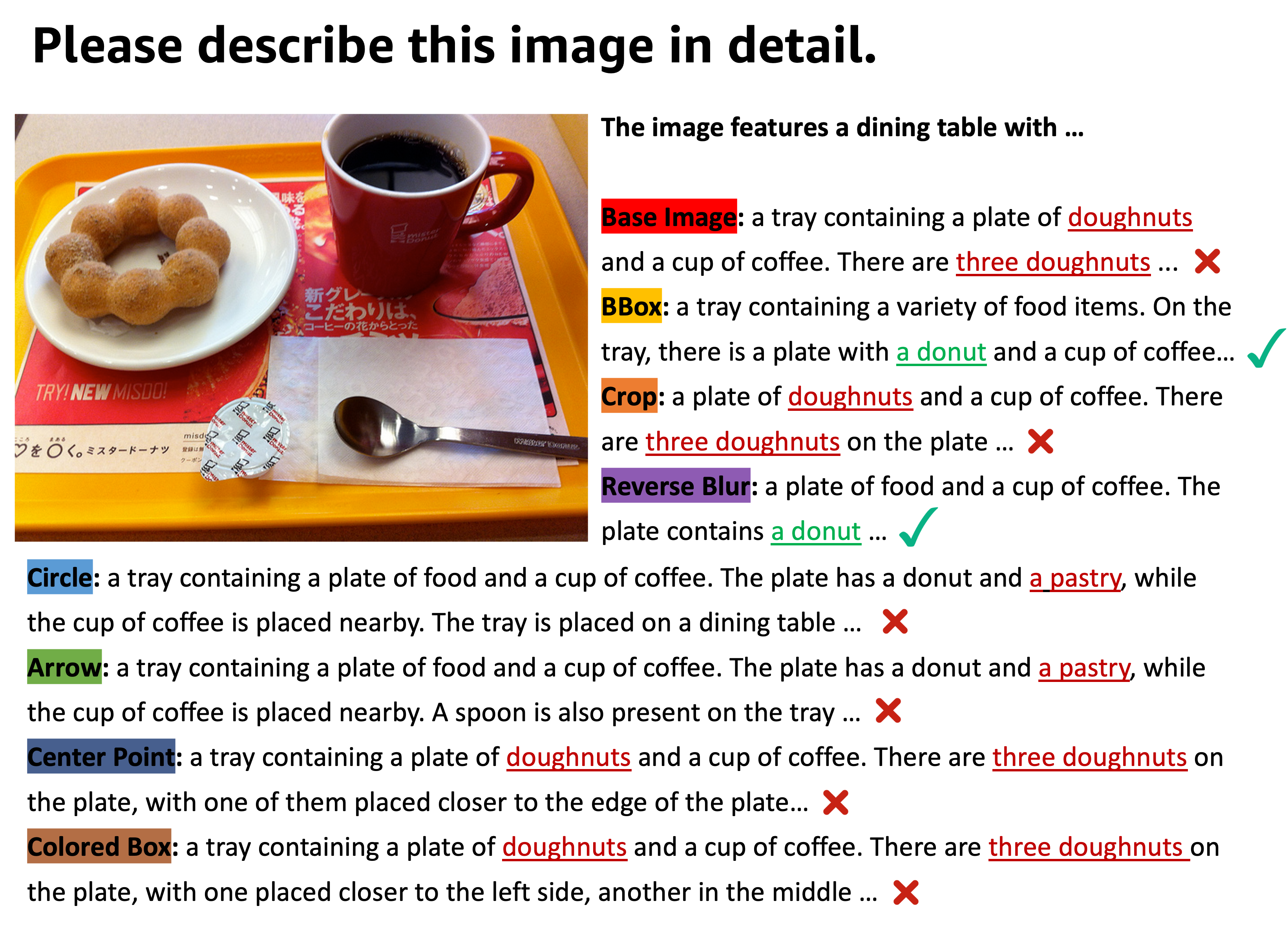}
    \vspace{-8mm}
    \caption{
        \textbf{Impact of different VPs on image description generation.}
        % Second question arises here.
        Different VPs produce varied results, but not all are equally effective.
        % Their success depends on the specific characteristics of the image.
        All responses are generated using greedy decoding to eliminate randomness and focus solely on the influence of visual prompting.
        % \todo{emphasize -- move fig 1}
    }
    % \kz{should we show the images with visual prompts applied, maybe more intuitive}}% \sm{included in figure 1}
    \vspace{-3mm}
    \label{fig:chair}
\end{figure}
\vspace{1mm}\noindent\textbf{Visual prompting for image description generation.}
\Cref{fig:chair} analyzes the impact of VPs on image descriptions.
While certain VPs, such as Bounding Box and Reverse Blur, enable the model to accurately identify existing items, others introduce errors by mentioning additional pastries or multiple donuts.
This again confirms the variability in VPs' effectiveness and underscores the importance of selecting the right VP to mitigate hallucination.
% \Cref{fig:chair} shows an analysis of various VPs for image description.
% It features a tray with a donut and a cup of coffee.
% Several visual prompts generate different interpretations, some of which contain errors.
% Some prompts, like Bounding Box and Reverse Blur, correctly identify the items, while others incorrectly mention multiple doughnuts or additional pastries.
% The analysis shows that VP effectiveness varies, with certain prompts being more accurate than others.

% 1. Why We Used Predefined VPs?
% From an optimization perspective, treating visual prompt engineering as discrete selection is cost-effective and straightforward (see L120–L131 of our paper).
% Even with only 8 predefined VPs, the Oracle achieves near-perfect performance, demonstrating the effectiveness of this approach.
% While additional VPs might provide finer granularity, our current setup delivers strong performance, especially with automated VP selection.

% Furthermore, one of our key contributions is the research finding that there is a near-optimal visual prompting method for each instance, highlighting the potential of this approach of overlaying visual prompts on images to mitigate object hallucination. We believe this finding will inspire future research to explore and further enhance performance in this direction. Our work lays a strong foundation for these future endeavors.
\section{Conclusion}
In this work, we proposed \textbf{BBVPE} framework to systematically identify optimal VPs that mitigate object hallucinations in LVLMs.
Our findings confirm that: \textbf{(A1)} carefully curated visual prompting can effectively reduce hallucinations in LVLMs, and \textbf{(A2)} optimal VPs can be systematically learned in a \textit{black-box} setup.
By dynamically selecting the most suitable VP from a predefined pool, guided by a trained router model based on LVLM preferences, our framework significantly enhances the performance of both open-source and proprietary LVLMs on hallucination benchmarks.
\section*{Limitations \& Future Work}
\noindent\textbf{(1)} Our current approach primarily focuses on natural images and does not extend to abstract and synthetic figures, such as those used in document VQA~\cite{mathew2021docvqa}, science VQA~\cite{lu2022learn}, or math VQA~\cite{lu2023mathvista}.
The current design of our method may not be directly applicable to these synthetic images, which typically exhibit different visual characteristics.

\noindent\textbf{(2)} We currently use bounding box-based prompts from the Segment Anything Model~\cite{kirillov2023segment}.
Transitioning to fine-grained, mask-based VPs could potentially enhance performance, as demonstrated in recent studies~\cite{yang2023set,yang2023fine}.

\noindent\textbf{(3)} Our router model currently considers only image features and does not incorporate the question context.
Our preliminary experiments suggest that incorporating question context could further improve results, pointing toward future work on exploring question-aware visual prompting.

% While current Oracle only considers image only, 
% Future work could focus on dynamically generating question-aware VPs.

% \noindent(4) To ease the optimization process and reduce the selection space, we have limited visual prompting to object-level granularity.
\noindent(\textbf{4)} To simplify optimization, we focus on object-level visual prompting, but extending to patch-based or pixel-based VPs could potentially provide a richer set of design space.

\noindent\textbf{(5)} Exploring the synergy between visual and textual prompt optimization remains an open research direction that may offer valuable insights.

\noindent\textbf{(6)} While our method is specifically designed to address object hallucination, exploring how VP and our framework perform in addressing attribute and relation hallucination remains an intriguing challenge that we leave for future work.

% \todo{Another area of interest is to investigate whether a single router model can generalize across different models, datasets, and evaluation benchmarks.}

\noindent\textbf{(7)} Object localization matters. We observed that better localization, such as using ground truth object coordinates, leads to improved results in our preliminary results.

\noindent\textbf{(8)} During router model training, we observed sensitivity to hyperparameters and occasional convergence instability, sometimes leading to overfitting.
This highlights the subtle learning signal from LVLM preferences over VPs, requiring a carefully designed training process.

% that convergence was not always stable and the model was somewhat sensitive to hyperparameters, which sometimes led to overfitting.
% This suggests that the learning signal from LVLMs' preference over VPs is subtle and challenging to capture effectively.
% A carefully designed training procedure is required to address these issues and prevent overfitting.

\noindent \textbf{Despite these limitations}, to the best of our knowledge, our study is the first black-box approach for mitigating object hallucination in LVLMs.
We hope that our initial investigation into automated visual prompt engineering and black-box strategies inspires further research into broader vision-language challenges beyond object hallucination.

\section*{Ethical Considerations}
In our current method, we use a predefined pool of VPs and have not observed any jail-breaking phenomena with visual prompting.
However, we are uncertain whether more fine-grained visual prompt engineering, such as using diffusion models, could lead to adversarial attacks or jail-breaking scenarios.
Rigorous testing is needed to ensure the robustness and safety of this approach.
Further research should address these considerations, if present, and focus on identifying and mitigating potential risks associated with VP misuse.

\bibliography{custom}
\newpage
\appendix

\twocolumn[{
\begin{center}
    \vspace*{\fill}
    \section*{\centering\LARGE Appendix}
    \vspace*{4mm}
    % \addcontentsline{toc}{section}{Appendix}
\end{center}
}]

\section{Implementation Details}
\label{sec:implementation_details}
% \noindent\textbf{Implementation details.}
We use a frozen CLIP-ViT-L/14@336px\footnote{\scriptsize\url{https://huggingface.co/openai/clip-vit-large-patch14-336}} model with a trainable MLP head as our VP router.
The router is trained on the COCO dataset~\cite{lin2014microsoft} training split, where each image is paired with 6 questions: 3 positive (about objects present in the image) and 3 negative (about objects not present in the image), following the POPE protocol~\cite{li2023evaluating}.
Each VP router is individually trained for each LVLM, as the preference for VPs varies across models, and we observed that these preferences do not transfer between models.
The training configuration is outlined below.
\begin{table}[h!]
    \centering
    \vspace{-3mm}
    \scalebox{1}{
    \begin{tabular}{y{70}|y{80}}
    config                   & value                     \\ \hline
    image size               & 336$\times$336                    \\
    optimizer                & AdamW                                \\
    learning rate            & 1e-4                              \\
    loss function            & cross entropy loss                \\
    training epochs          & 20                          \\
    \end{tabular}
    }
    \vspace{-3mm}
    \caption{\textbf{Training configurations for the router model.}}
    \vspace{-3mm}
\end{table}

\noindent For the object localizer, we use Segment Anything Model 2 (\texttt{sam2-hiera-large})\footnote{\scriptsize\url{https://huggingface.co/facebook/sam2-hiera-large}}.
For LVLMs, we use two open-source models, LLaVA-1.5-7b\footnote{\scriptsize\url{https://huggingface.co/liuhaotian/llava-v1.5-7b}} and InstructBLIP-vicuna-7b\footnote{\scriptsize\url{https://huggingface.co/Salesforce/instructblip-vicuna-7b}}, and two proprietary
models, GPT-4o (\texttt{gpt-4o-2024-08-06})\footnote{\scriptsize\url{https://platform.openai.com/docs/models}} and Claude-3.0-Sonnet (\texttt{claude-3-sonnet-20240229})\footnote{\scriptsize\url{https://docs.anthropic.com/en/docs/about-claude/models}}.

\section{More Details on Evaluation Setup}
\label{sec:evaluation_details}
\vspace{1mm}\noindent\textbf{Benchmarks.}
% We evaluate object hallucinations in LVLMs using discriminative and descriptive tasks.
% In discriminative tasks, models answer specific questions about object presence --- evaluated using the POPE benchmark.
% For descriptive tasks, models generate image descriptions --- assessed using the CHAIR benchmark.
We evaluate object hallucinations in LVLMs through discriminative and descriptive tasks on the COCO~\cite{lin2014microsoft} validation split, using the POPE and CHAIR benchmarks, respectively.

\vspace{1mm}\noindent\textbf{(1) POPE}~\cite{li2023evaluating} frames hallucination assessment as a binary classification task, asking yes/no questions about the presence of both real and nonexistent objects in an image (\eg, “\textit{Is there a/an \texttt{[OBJECT]} in the image?}”).
Questions for real objects are randomly selected from the actual objects present in the image.
There are three prompt setups for selecting nonexistent objects:
\begin{itemize}[leftmargin=*, labelsep=0.3em]
    \vspace{-2mm}\item Random: Nonexistent objects are randomly selected from all object categories.
    \vspace{-3mm}\item Popular: Nonexistent objects are chosen from top-$k$ most frequent objects in the dataset.
    \vspace{-3mm}\item Adversarial: Objects are chosen based on frequent co-occurrences with actual objects but are absent from the image.
    % Objects are ranked based on their co-occurrence frequencies with actual objects in the image. Nonexistent objects are selected from the top-$k$ frequently co-occurring but absent objects.
    \vspace{-2mm}
\end{itemize}
We use Accuracy, Precision, Recall, and F1 score as evaluation metrics.
Accuracy reflects the proportion of correctly answered questions.
Precision and Recall indicate the correctness of “Yes” and “No” answers, respectively.
F1 score is a harmonic mean of Precision and Recall.

\vspace{1mm}\noindent\textbf{(2) CHAIR}~\cite{rohrbach2018object} 
evaluates the proportion of words in captions that correspond to actual objects in an image, based on ground-truth captions and object annotations.
% This metric measures how accurately the generated words align with the objects in the image, according to the ground truth.
The metric has two variants:
\begin{itemize}[leftmargin=*, labelsep=0.3em]
    \vspace{-2mm}\item Per-sentence (CH$_S$): Proportion of sentences containing hallucinated objects, calculated as CH$_S$ = $\frac{|\text{{\# sentences with hallucinated objects}}|}{|\text{{\# all sentences}}|}$.
    \vspace{-2mm}\item Per-instance (CH$_I$): Proportion of hallucinated objects relative to all mentioned objects, calculated as CH$_I$ = $\frac{|\text{{\# hallucinated objects}}|}{|\text{{\# all objects mentioned}}|}$.
    \vspace{-2mm}
\end{itemize}
% We perform image captioning with the prompt, “\textit{Please describe this image in detail.}” to generate captions for evaluation.
Captions are generated with the prompt, “\textit{Please describe this image in detail.}” for evaluation.

% \vspace{1mm}\noindent\textbf{Instruction for GPT-4o evaluation.}
% \cref{fig:gpt4o_instruction} shows the instruction given to GPT-4o for evaluating textual descriptions of an image, based on five criteria: Accuracy, Detail, Comprehensiveness, Relevance, and Robustness.
% Each criterion is scored on a scale from 1 to 10, with higher scores reflecting better performance.

\section{Instruction for GPT-4o Evaluation}
\label{sec:gpt4o_evaluation}
\cref{fig:gpt4o_instruction} shows the instruction given to GPT-4o for evaluating 8 textual image descriptions of an image, based on 5 criteria: Accuracy, Detail, Comprehensiveness, Relevance, and Robustness.
Each criterion is scored on a scale from 1 to 10, with higher scores reflecting better performance.
Total scores are calculated for each description to evaluate their overall quality.

\clearpage
\begin{figure*}[p!]
    \centering % Center the box
    \begin{tcolorbox}[
        title=Image Description Quality Assessment using GPT-4o, 
        colframe=blue!50!black, 
        colback=blue!5!white, 
        coltitle=white, 
        fonttitle=\bfseries,
        width=\textwidth,
        breakable
    ]
    \begin{adjustbox}{minipage=\textwidth,scale=0.85}
    \begin{verbatim}
<SYSTEM_MESSAGE>
You are an expert in image description evaluation. Your task is to assess how well textual 
descriptions capture the detailed visual information of images.

<INSTRUCTION>
Compare and evaluate the following 8 descriptions of the provided image.

Descriptions:
{description 1}
{description 2}
...
{description 7}
{description 8}

For each description, rate a score on a scale of 1 to 10, where a higher score indicates better 
performance, for each of the 5 criteria:
1. Accuracy: How precisely does the description reflect the actual objects, details, and 
attributes (such as color, shape, and number of objects) visible in the image?
2. Detail: How thoroughly does the description capture visual details of the objects, including 
finer elements like positions, relative sizes, and relationships?
3. Comprehensiveness: How well does the description cover all key elements of the image, without 
omitting important objects or details?
4. Relevance: Does the description focus on significant and pertinent details from the image. The 
score decreases if the description includes unnecessary or unrelated information that distracts 
from the core details of the image.
5. Robustness: Does the description avoid mentioning any objects or attributes that are not 
present in the image? Descriptions without any false information score higher. If nonexistent 
elements are included, the score decreases.

Only provide the numerical scores for each criterion and the total score, formatted as follows:
1. Accuracy: score1 | score2 | score3 | score4 | score5 | score6 | score7 | score8
2. Detail: score1 | score2 | score3 | score4 | score5 | score6 | score7 | score8
3. Comprehensiveness: score1 | score2 | score3 | score4 | score5 | score6 | score7 | score8
4. Relevance: score1 | score2 | score3 | score4 | score5 | score6 | score7 | score8
5. Robustness: score1 | score2 | score3 | score4 | score5 | score6 | score7 | score8
Total Score: total1 | total2 | total3 | total4 | total5 | total6 | total7 | total8
    \end{verbatim}
    \end{adjustbox}
    \end{tcolorbox}
    \caption{
        \textbf{GPT-4o evaluation instruction.}
    }
    \label{fig:gpt4o_instruction}
\end{figure*}

% \section{Details on GPT-4o Evaluation}
% \label{sec:gpt4o_details}
% \paragraph{User prompt.}
% \todo{placeholder placeholder placeholder placeholder placeholder placeholder placeholder placeholder placeholder placeholder placeholder placeholder placeholder}

% \section{Qualitative Results}
% \label{sec:qualitative_results}
% POPE baseline vs. ours
% CHAIR baseline vs. ours

% \section{Tips and Tricks}
% \label{sec:tips_and_tricks}
% \paragraph{Textual prompt.}
% Simple trick to enhance the performance
% \begin{tcolorbox}[colback=gray!5!white, colframe=gray!75!black, title=Single word constraint]
% "Is there a/an \{OBJECT\} in the image? \textcolor{red}{Answer using only a single word (Yes or No)}."
% \end{tcolorbox}
% Also mentioned in LLaVA~\cite{liu2023improved}.
% The addition of this simple prompt () significantly enhances Recall.

% \todo{additional prompt}

\end{document}